\documentclass[letterpaper]{article} 
\usepackage{aaai23}  
\usepackage{times}  
\usepackage{helvet}  
\usepackage{courier}  
\usepackage[hyphens]{url}  
\usepackage{graphicx} 
\usepackage[square,sort,comma,numbers]{natbib}  
\usepackage{caption} 
\usepackage{algorithm}
\usepackage{algorithmic}
\usepackage{amsmath}
\usepackage{multirow,hhline,array}
\usepackage{comment}
\usepackage{color}
\usepackage{amssymb}
\usepackage{adjustbox}
\usepackage{newfloat}
\usepackage{listings}
\usepackage{kotex}
\usepackage{titlesec}
\DeclareCaptionStyle{ruled}{labelfont=normalfont,labelsep=colon,strut=off} 
\floatstyle{ruled}
\newfloat{listing}{tb}{lst}{}
\floatname{listing}{Listing}
\begin{document}
%
\title{Semantics-Guided Object Removal for Facial Images:\\ with Broad Applicability and Robust Style Preservation}
\author{Jookyung Song, 
    Yeonjin Chang, 
    SeongUk Park,
    Nojun Kwak\\
    Seoul National University\\
    \{chsjk9005, yjean8315, swpark0703,
    nojunk\}@snu.ac.kr
}
\maketitle
\begin{abstract}
Object removal and image inpainting in facial images is a task in which objects that occlude a facial image are specifically targeted, removed, and replaced by a properly reconstructed facial image. Two different approaches utilizing U-net and modulated generator respectively have been widely endorsed for this task for their unique advantages but notwithstanding each method's innate disadvantages. U-net, a conventional approach for conditional GANs, retains fine details of unmasked regions but the style of the reconstructed image is inconsistent with the rest of the original image and only works robustly when the size of the occluding object is small enough. In contrast, the modulated generative approach can deal with a larger occluded area in an image and provides {a} more consistent style, yet it usually misses out on most of the detailed features. This trade-off between these two models necessitates an invention of a model that can be applied to any size of mask while maintaining a consistent style and preserving minute details of facial features. Here, we propose Semantics-Guided Inpainting Network (SGIN) which itself is a modification of the modulated generator, aiming to take advantage of its advanced generative capability and preserve the high-fidelity details of the original image.
By using the guidance of a semantic map, our model is capable of manipulating facial features which grants direction to the one-to-many problem for further practicability.
\end{abstract}

\section{Introduction}

Object removal and image inpainting in facial images is a task of removing objects that block the foreground human facial area and reconstructing the occluded facial features. As the occluding objects often have irregular shapes and sizes, and the hidden facial images are deprived of the semantics from the original image, the challenge that this task poses is not inconsequential. 
Generally, this task is comprised of three distinct challenges: 1) generalization in novel masks, 2) style consistency, and 3) preservation of known pixels. Object remover modules should be able to specifically erase any occluding objects regardless of their types, shapes, and sizes. The reconstructed image must maintain the style consistent with the unoccluded regions, while at the same time details of unoccluded regions should be preserved in high pixel-by-pixel fidelity.

\begin{figure}[t]
\centering
\includegraphics[width=\linewidth]{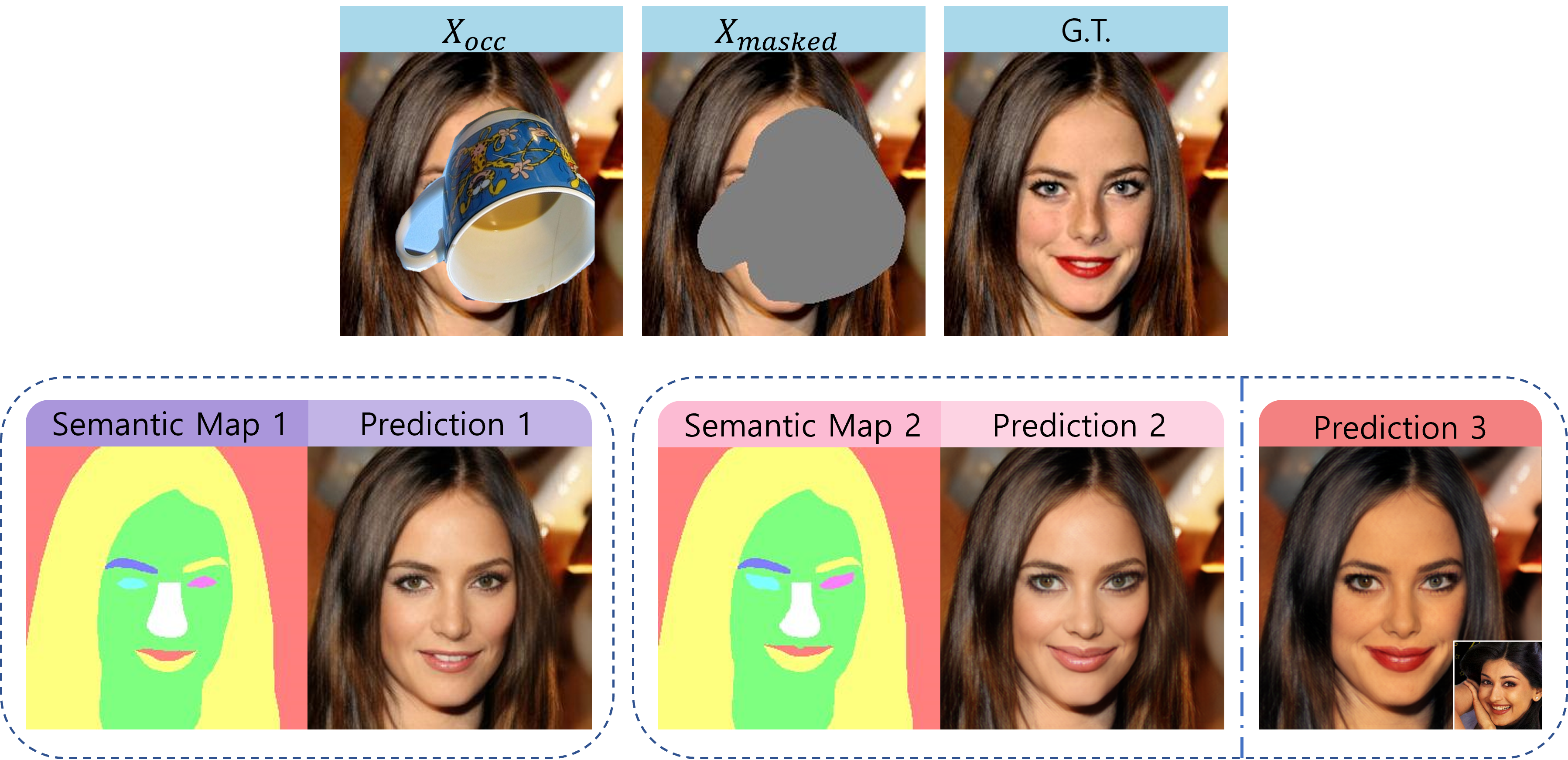}
\caption{Our method can provide various predictions by manipulating the semantic map, or the style latent vector. Prediction 3 is the example of referencing the style from right below image when using the semantic map 2.}
\label{fig:failure}
\end{figure}

Conditional generative adversarial networks (GANs) \cite{mirza2014conditional}, widely used in other generative tasks, have been the primary choice for this line of tasks. Conditional GANs used in the facial de-occlusion and reconstruction are classified into two categories with regard to their architecture for the generator part; U-net-based generator and modulated-generator-based approach. U-net-based generator focuses on completing only the masked region, and conventionally the rest {part of an} image is directly copy-and-pasted from the conditioned image. However, the fact that it does not train to construct the whole image is postulated to have {a} negative effect on its generative capability, as when {U-net-based generators are} confronted with novel masks {in shape and size} not seen during training, {they} tend to significantly underperform. Figure \ref{fig:qualitative} shows the failure cases of the U-net-based generator when tested on the different mask types unseen during the training. 


{Modulated generative approach is one of the recent advances among the conditional GANs, which regards an input as a random constant and each convolution layer adjust{s} the intermediate latent vectors with denormalization factors (e.g. scale and bias) \cite{karras2019style}. Modulated approach {has} further advanced the generative capability and edit{-}ability. However, in the case of conditional generation, there is information loss when a conditioned image is compressed to a low-dimensional latent vector. \cite{shannon1959coding} The lost information is mostly high-frequency details or the infrequent information such as background, as GAN tends to sustain common information of a domain. This leads to the model's under-performance at pixel preservation of the unmasked region as well, resulting in the prediction 
{being} largely different from the conditioned input. }

Due to these trade-offs, conventional object removal tasks are limited to certain types and shape{s} of occluding objects, limiting their applicability for real-world data. We aimed to construct a generator that can simultaneously deal with the aforementioned challenges. We introduced a number of changes to the modulated generative approach, creating high fidelity output{s} by utilizing additional semantic knowledge about the foreground, which is the semantic label map. The feat of this paper is the spontaneous generation of {the} semantic map without the need for {an} additional ground truth semantic map. In addition, the users have controllability over the output by designating the semantic map or style to their demands. Our primary contributions are as {follows}: 

\begin{itemize}
    \item  {With the help of semantic map predictor, our model provides the modulated generator with additional semantic knowledge
    {that embodies user's intention for image inpainting which is a well-known one-to-many problem.}} 
    \item {For the restoration of high-frequency information at the unmasked region, we adopt self-distillation loss and {the} fusion feedback network. }
    \item Our work provides a broad range of control over the disentangled semantic region, enabling various image manipulation including semantic manipulation and style swapping.
\end{itemize}

\begin{figure*}[ht]
\centering
\includegraphics[width=0.9\linewidth]{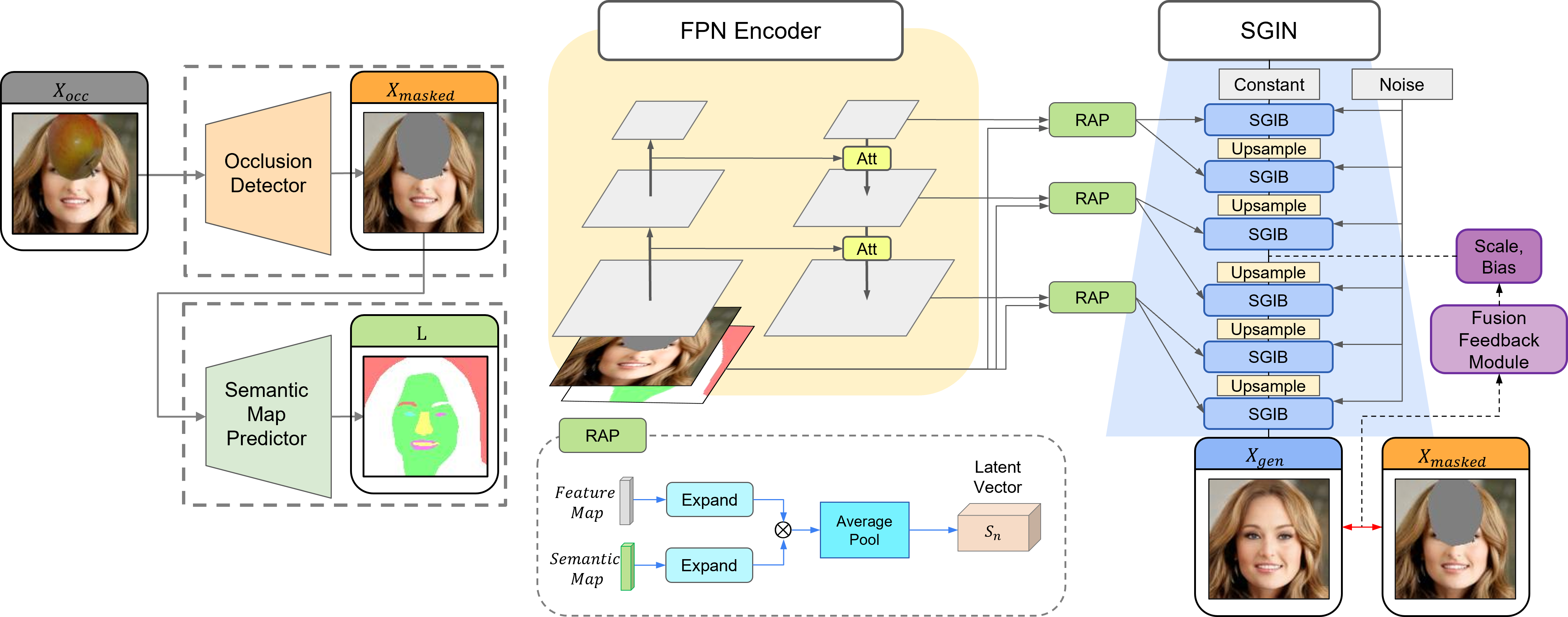}
\caption{\textbf{Overall SGIN architecture} 
The masked image $X_{masked}$ and the semantic predicted map $L$ are concatenated and passed through {the} FPN encoder. \textit{Attn} denotes the contextual attention module. Each encoded feature map is processed through {the} \textit{RAP} block to produce the GAP-ed semantic style latent vectors {$S = \{S_1, \cdots, S_N\}$}. Each vector $S_n$ is passed to \textit{SGIB} and the generator produces the coarse prediction $X_{gen}$. The $L_2$ difference between $X_{gen}$ and $X_{masked}$ is combined through {the} \textit{Fusion Feedback Module}, and is backpropagated into the middle-most \textit{SGIB} layer.}
\label{fig:overall}
\end{figure*}

\begin{figure}[ht]
\centering
\includegraphics[width=1.\linewidth]{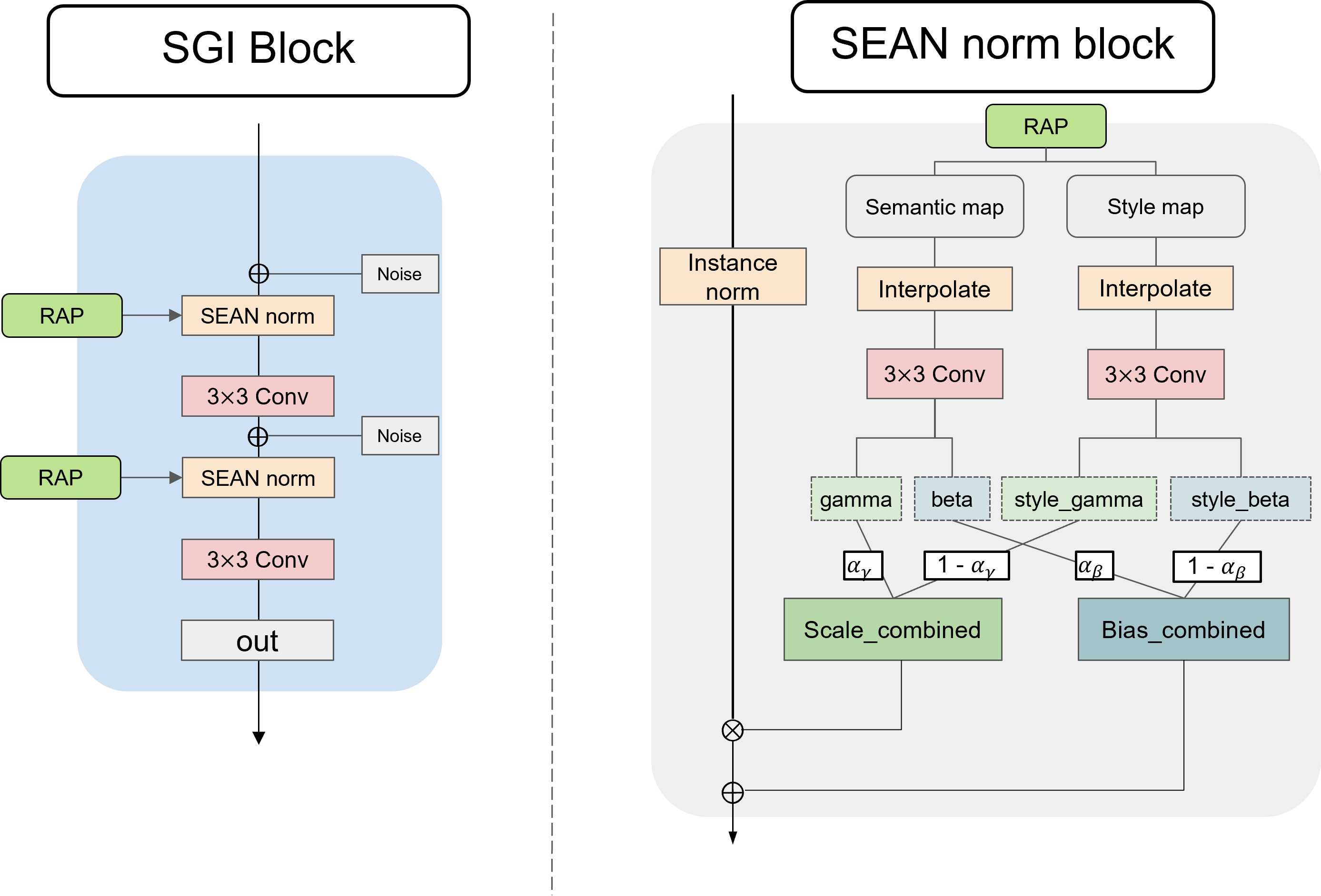}
\caption{\textbf{SGI block and Normalization block}} 
\label{fig:SGIB}
\end{figure}

\section{Related Work}
\noindent\textbf{Image Inpainting}
{is defined as a task of reconstructing missing regions in an image as well as removing objects that {occlude an} image.}
Early non-learning{-based} approaches {used to fill} 
the masked regions with {the} information {retrieved directly from their surrounds} \cite{bertalmio2000image, telea2004image}, or replace the missing regions with the {best matching patch} 
\cite{efros1999texture, efros2001image, criminisi2003object, criminisi2004region, barnes2009patchmatch}.
With the {advent} of deep learning, various learning-based methods for image inpainting have been studied. 
\cite{pathak2016context} was the first to use {GANs} \cite{goodfellow2014generative} for image inpainting, using an encoder-decoder architecture.
{Until now,}
numerous studies using the U-Net architecture have been conducted \cite{iizuka2017globally, li2017generative, yan2018shift, zeng2019learning} 
{and more recently,} the modulated generator approaches started to generate photo-realistic images, while some studies utilized them as generators and trained encoders correspond to them \cite{richardson2021encoding, tov2021designing}.


\noindent\textbf{Semantic Image Inpainting}
{refers to the problem of filling in large holes that require semantic information. 
Even with the advances in GANs, image inpainting is still an ill-posed problem, especially when most of the semantic knowledge is lost. }
There have been a number of novel approaches to the problem \cite{yeh2017semantic,li2017generative}. 
\cite{liu2018image} pointed out that existing studies have focused only on rectangular-shaped holes, and proposed PConv to address irregular masks by constructing the U-Net architecture with partial convolution layers.
\cite{yu2018generative} proposed a contextual attention layer to explicitly utilize surrounding image features, overcoming the ineffectiveness of convolutional neural networks in explicitly utilizing distant surrounding information.

As semantic image inpainting is an ill-posed problem, several attempts have been made to utilize additional information. {EdgeConnect \cite{nazeri2019edgeconnect} is a two-stage adversarial network where an edge generator generates surrounding edges in an image and an image completion module is provided with this addition{al} information about the edges. There has also been an attempt to integrate landmark information of the face \cite{xinyi2021identity} or utilize the semantic maps for facial reconstruction. \cite{liao2018face, hong2018learning}}




\section{Method}
Our method {consists} of three different modules; 1) Occlusion Detector 2) Semantic Map Predictor 3) {Semantics-Guided Image Inpainter (SGIN)}. 
First, the \textit{occlusion detector} detects the occluding region in the $X_{occ}$  {and we mask the occluding region in $X_{occ}$ to obtain $X_{masked}$.}
Then, the \textit{semantic map predictor} predicts the semantic map of the $X_{masked}$. {Finally, \textit{SGIN} properly inpaints the masked area.}
{Fig.~\ref{fig:overall}} illustrates the overall architecture of our framework and we describe the details of each module {below}.

\subsection{Occlusion Detector} 
Given the input $X_{occ}$, {our occlusion detector} detects the occluding object{s} and predicts {the} binary mask $M$ {(1: occlusion / 0: non-occlusion)}. The masked input is denoted as $X_{masked} = X_{occ} \odot(\mathbf{1}- M)$ {where $\mathbf{1}$ is an all-one image and $\odot$ denotes element-wise product}. We built the network upon {a simple} ResNet {generator} architecture constructed with five-layers of convolution blocks and five-layers {of} deconvolution blocks. 

\subsection{Semantic Map Predictor} 
{The usage of semantic map allows us to grasp both the purpose and performance of the model.}
Although requiring the semantic labeling can {accompany much efforts}, we overcome this by using a semantic map predictor which enables obtaining the semantic label in {an} on-the-fly manner so that we can neglect the need of human labeling.
It is important to note that the \textit{semantic map predictor} is a pre-trained network trained with a {separate} non-overlapping dataset with the SGIN's training data, and fortunately it generalizes well to the SGIN's training data.
Given the masked image $X_{masked}$, the semantic map predictor predicts the semantic label map {$L = \{l_{1}, \cdots, l_{C} \}$. Each $l_c , c \in [C]$,} indicates {the} binary class label map for eleven regions {(i.e, $C=11$)}. We trained BiseNet \cite{yubilateral} for {our} generalized semantic map predictor. 

\subsection{Semantic Style {Encoding}}
We chose Feature Pyramid Network (FPN) \cite{richardson2021encoding} as our encoder, which generates latent codes through multi-scaled hierarchical features. 
Style representations from the latent code are fully determined by the masked image $X_{masked}$ and the semantic label map {$L_{n}, n \in[N]$, where $N$} indicates the number of layers in the FPN's feature map.
As we use the semantic map $L_{n}$ for additional semantic knowledge, the output latent code needs to {disentangle 
styles {(e.g, color, patterns ...)} for each semantic region}. 
In order to achieve this, we first concatenate the semantic label $L_{n}$ and the masked image $X_{masked}$ channel-wise. 
Then each pyramid network produces $F_{n}$, where $H_{n}$ {and} $W_{n}$ indicate the spatial dimension of height and width for each layer. 
We expand $F_{n}\in \mathbb{R}^{H_{n} \times W_{n} \times 512}$ to $F_{exp_{n}} \in \mathbb{R}^{H_{n} \times W_{n} \times 512 \times C}$  by broadcasting {along} the dimension of binary class map. 
Also, the semantic {label map} $L_{n} \in \mathbb{R}^{H_{n} \times W_{n} \times C}$ is broadcasted {along the} dimension of feature map channels, producing $L_{exp_{n}} \in \mathbb{R}^{H_{n} \times W_{n} \times 512 \times C}$. 
Additionally, it is difficult to construct faithful style embeddings in the missing holes, because there are no features extracted from the masked region. In the light of this, we harness the well-known contextual attention module \cite{yu2018generative} in between the feature pyramids, which can provide additional attention-wise information in the masked region as well. Ablation study shows that {the} attention module helps increasing the prediction quality.

\subsection{Region-wise Average Pooling {(RAP)}}
As the label map is binary, we can extract semantic latent codes which contain activations for each semantic {region} by multiplying {the latent code with the semantic label map}. 
Each activation is spatially {global-}average-pooled (GAP), so the outcome latent space {is} {$S = \{S_1,\cdots,S_N\}$}, where $S_n = \text{GAP} (F_{exp_n}\odot L_{exp_n})$. 

\subsection{Semantics-Guided Inpainting Network (SGIN)}
The semantics{-}guided inpainting network (SGIN) is comprised of a number of serially connected SGI blocks (SGIB), {the exact number of which} is determined by the resolution of training images. {As shown in Fig. \ref{fig:SGIB}, each} convolution block has two convolution layers, which is {composed of} a normalization layer and {a} convolution layer. The input in the normalization layer is first{ly} instance{-}normalized, and then denormalized by the semantic region adapative block (SEAN) \cite{zhu2020sean} in an attempt to reflect the previously extracted semantic features on the reconstructed image. 
While the spatially adaptive (SPADE) normalization block \cite{park2019semantic}, which can separately process spatial parameters of each image, we chose SEAN, a variant of SPADE, as our denormalizer as it is able to process not only the spatial parameters but also style modulation parameters.
Overall, the generator of our framework is expressed as follows;
\begin{equation}
    G(X_{masked},L) = \text{SGIN}(\text{RAP}(\text{FPN}(X;L));L)
\end{equation}
After passing through the convolution blocks, the `coarse' output data is compared with the original images overlaid by a mask (conditioned images) and the differences in the corresponding pixels are denoted in the form of MSE loss.

\subsection{Fusion Feedback Module}

As mentioned previously, the latent vectors of the modulation generative approach tend to lose high-frequency details because of their lossy data compression.
Making up for this loss is the key to best achieving our method's goal towards high fidelity generation. 

Feeding back the lost features directly to the generation module has been one of the best solutions to this problem \cite{wang2022high}. Yet, such a model requires heavy memory with large computational units such as consultation fusion mapping networks, adaptive distortion mapping networks and various data augmentation techniques.

To tackle these problems, we introduce {the} Fusion Feedback Module, which uses a {lightweight} encoder-decoder network that compares pixel values of the initially reconstructed image with those of the original masked image from the input to retrieve lost details, without any need for additional modules. After the generator draws its first `coarse image', which is deprived of finer features of the original image, the $L_2$ difference between the generated image and the ground truth is calculated, and this crude generated image along with $L_2$  difference is injected to the middlemost layer of the generator module, which is experimentally shown to generate images with the highest fidelity.

\subsection{Loss Function}
We introduce the concept of self-distillation loss, which provides the feature-level supervision directly to the generator for preserving high-fidelity details of the input.
Inspired by the `privileged information' in the work of PISR \cite{lee2020learning}, {where} a teacher network is forwarded with a ground truth image to produce further detailed features and a student network learns the feature map of the teacher network through distillation, we devised an information flow that the generator is fed with its own first coarse image along with the loss calculated from the comparison between the feature map of the ground truth and the predicted output. {Thus, we call this as} the \textit{self-distillation loss}. The details of the calculation are as follows:
The generator is forwarded with the ground truth image $X_{gt}$ with no masked region, and produce compact feature maps $f_{i}(X_{gt})$ in {the} $i^{th}$ SGI {block}.
Then, the $L_2$ difference between the feature map of the initially forwarded masked input $X_m$ and the {groundtruth,} $X_{gt}$, is calculated.
Finally, the self distillation loss {is defined as $\mathcal{L}_{sd} = \sum_{i=1}^{K} ||f_{i}(X_{gt}) - f_{i}(X_m)||_2$, where $K$} denotes the number of SGI blocks. 
The advantageous effect of using self-distillation loss can be found in the ablation study section.

In addition, we applied several conventionally used loss functions in the {literature} of image inpainting. The discriminator computes the $\mathcal{L}_{feat}$, which is the $L_1$ loss between the discriminator features for the $X_{gt}$ and the predicted image, as well as an adversarial loss $\mathcal{L}_{adv}$. Also, we used the $\mathcal{L}_{per}$, which is the perceptual loss between the features of $X_{gt}$ and $X_{masked}$ extracted from a VGG-19 network \cite{simonyan2014very}. {$\mathcal{L}_{adv}$ and $\mathcal{L}_{per}$ are defined as follows}:         
\begin{equation}
    \mathcal{L}_{adv} = \mathbb{E}_X[\log D(X))] + \mathbb{E}_X[\log (1-D(G(X_{masked} |L))],
\end{equation}
\begin{equation}
    \mathcal{L}_{per} = ||\text{Vgg}(G(X_{masked} |L)) - \text{Vgg}(X)||_2.
\end{equation}
Here, $G(\cdot|L)$ is the generator conditioned on the semantic map $L$, {while} $D(\cdot)$ denotes the discriminator.
The {overall loss} is as {follows}:
\begin{equation}
    \mathcal{L} = \lambda_{sd}\mathcal{L}_{sd} +  \lambda_{feat}\mathcal{L}_{feat} + 
    \lambda_{per}\mathcal{L}_{per} +
    \lambda_{adv}\mathcal{L}_{adv} .
\end{equation}
For the loss weights {$\lambda$'s}, please {refer to} the supplementary details. 
\section{Experiments}
\subsection{Face Occlusion Datasets}
\begin{figure}[t]
\centering
\includegraphics[width=1.\linewidth]{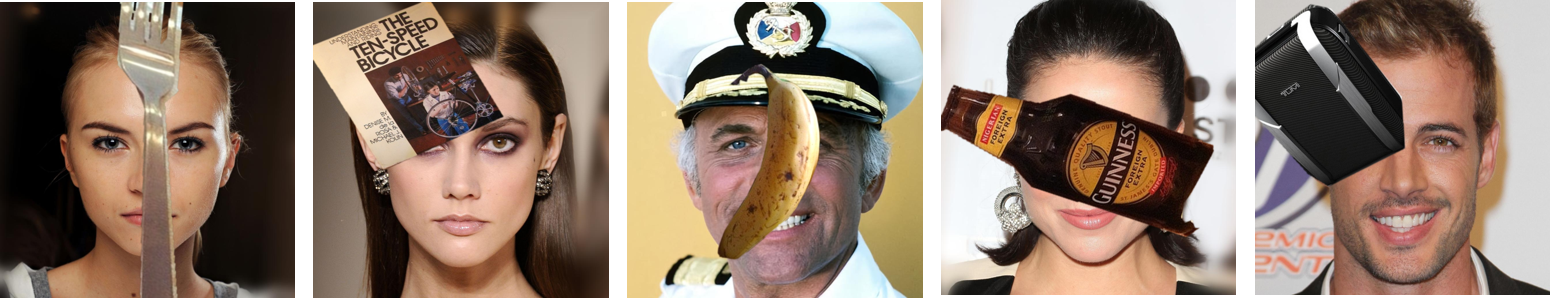}
\caption{\textbf{Examples of \textit{NatOcc} on \textit{CelebA-HQ} dataset}}
\label{fig:NatOcc}
\end{figure}

To generate diverse occluded facial images, we used \textit{Naturalistic Occlusion Generation (NatOcc)} \cite{voo2022delving} to overlay human facial images from HELEN \cite{le2012interactive} and CelebA-HQ \cite{lee2020maskgan} with occluding objects and create naturalistic synthetic images {(See Fig.~\ref{fig:NatOcc} for some examples).} As for the occluding objects, {we} used 128 objects across 20 categories from Microsoft Common Objects in Context (COCO) and 200 hands from EgoHands \cite{bambach2015lending}. Note that for the training of {our} semantic map predictor, {HELEN-derived occlusion images are used and its evaluation is done using CelebA-HQ images}. {Different from this,} for the SGIN, we only used CelebA-HQ. We split CelebAMask-HQ-derived images into 22,300 training images and 2,800 validation images. 
For more implementation details, please refer to the supplementary information. 

\begin{figure*}[h!]
\centering
\includegraphics[width=1.\linewidth]{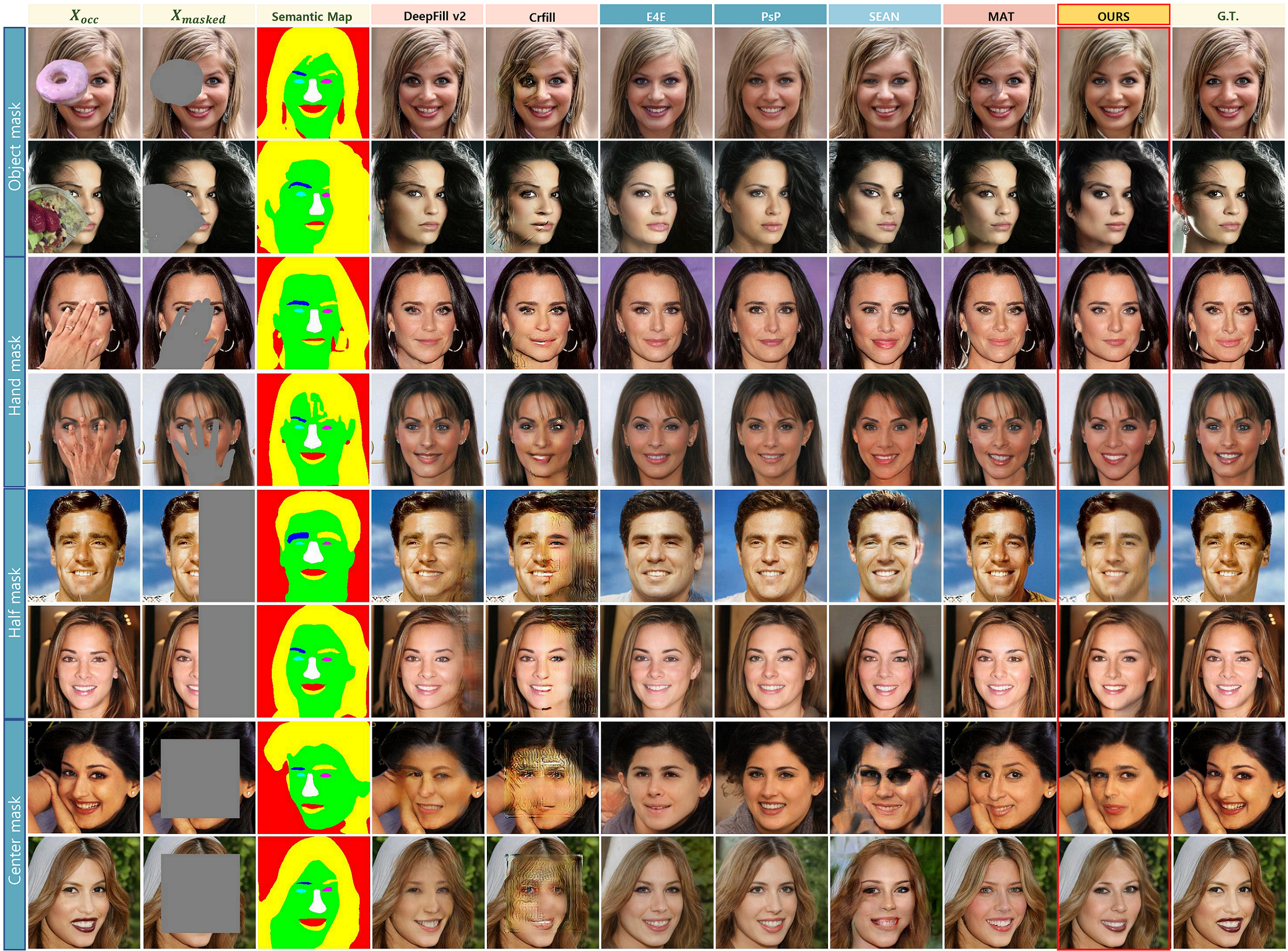}
\caption{\textbf{Qualitative Analysis of seven inpainting models.}}
\label{fig:qualitative}
\end{figure*}

\begin{table*}[h!]
\centering
\adjustbox{max width=\textwidth}{
\begin{tabular}{c|cc|cc|cc|cc|cc|cc|cc}
\noalign{\smallskip}\noalign{\smallskip}\hline\hline
\multirow{2}{*}{models} & \multicolumn{4}{c|}{U-net} & \multicolumn{4}{c|}{Modulated generator} & \multicolumn{2}{c|}{Semantic Map} &
\multicolumn{2}{c|}{Transformer} &
\multicolumn{2}{c}{Ours}\\
\cline{2-15}
      & \multicolumn{2}{c|}{DeepFill-v2}  & \multicolumn{2}{c|}{Crfill} & \multicolumn{2}{c|}{PsP} & \multicolumn{2}{c|}{E4E} & \multicolumn{2}{c|}{SEAN} & \multicolumn{2}{c|}{MAT} & \multicolumn{2}{c}{\textbf{SGIN (ours)}} \\
\hline
 PSNR(+) & 25.34 & \textcolor{red}{25.56} & 25.52 & \textcolor{red}{25.52}   & \textcolor{blue}{18.58} & 26.57   & \textcolor{blue}{19.36} & 27.13   & \textcolor{blue}{16.96} & 24.79   & 25.04 & \textcolor{red}{25.23}   & \textcolor{blue}{\textbf{23.76}} & \textcolor{red}{\textbf{26.62}}   \\
 SSIM(+)   & 0.9 & \textcolor{red}{0.91}   & 0.90 & \textcolor{red}{0.90}   & \textcolor{blue}{0.54} & 0.92   & \textcolor{blue}{0.56} & 0.92   & \textcolor{blue}{0.51} & 0.90   & 0.87 & \textcolor{red}{0.90}   & \textcolor{blue}{\textbf{0.74}} & \textcolor{red}{\textbf{0.92}}  \\
 MS-SSIM(+)   & 0.90 & \textcolor{red}{0.88}   & 0.91 & \textcolor{red}{0.90}   & \textcolor{blue}{0.66} & 0.94   & \textcolor{blue}{0.69} & 0.91   & \textcolor{blue}{0.63} & 0.87   & 0.91 & \textcolor{red}{\textbf{0.91}}   & \textcolor{blue}{\textbf{0.87}} & \textcolor{red}{0.90}  \\
 RMSE(-)   & 14.71 & \textcolor{red}{14.71}   & 14.45 & \textcolor{red}{14.45}   & \textcolor{blue}{30.80} & 12.85   & \textcolor{blue}{28.19} & 12.10   & \textcolor{blue}{37.38} & 15.89   & 15.38 & \textcolor{red}{15.38}   & \textcolor{blue}{\textbf{17.18}} & \textcolor{red}{\textbf{12.94}}    \\

LPIPS(-)   & 0.06 & \textcolor{red}{0.06}   & 0.08 & \textcolor{red}{0.08}   & \textcolor{blue}{0.23} & 0.04   & \textcolor{blue}{0.20} & 0.05   & \textcolor{blue}{0.26} & 0.06   & 0.06 & \textcolor{red}{\textbf{0.05}}   & \textcolor{blue}{\textbf{0.12}} & \textcolor{red}{\textbf{0.05}}  \\
 FID(-)   & 5.00 & \textcolor{red}{17.03}   & 9.95 & \textcolor{red}{28.83}   & \textcolor{blue}{38.50} & 6.31   & \textcolor{blue}{25.98} & 6.59   & \textcolor{blue}{25.77} & 6.88   & 4.21 & \textcolor{red}{14.27}   & \textcolor{blue}{\textbf{8.55}} & \textcolor{red}{\textbf{8.25}}  \\
\hline
\hline
\end{tabular}}
\caption{Quantitative comparison on CelebA-HQ with $256 \times 256$ resolution. 
The table shows two different scores for each models, the left one measured with full images and the right one measured with only the masked region. The color red designates proper scores for models which copy-and-paste the unmasked region and the color blue designates proper scores for models which generate the whole image. It is appropriate to compare the score{s} {in the} same color.}
\label{tab:model_compare}
\end{table*}

\begin{table*}[ht]
\begin{center}

\adjustbox{max width=\textwidth}{
\begin{tabular}{c|ccc|c|c}
\noalign{\smallskip}\noalign{\smallskip}\hline\hline
metrics & \multirow{1}{*}{w/o fusion feedback} & \multirow{1}{*}{w/o attention} & \multirow{1}{*}{w/o self-distill loss} 
 & \multirow{1}{*}{\textbf{FULL}} & \multirow{1}{*}{G.T. semantics}\\
\hline
 PSNR(+) & 19.181 (-4.544) & 21.844 (-1.881) & 22.395 (-1.33) & 23.725 & 24.051    \\
 SSIM(+) & 0.546 (-0.212) & 0.767 (-0.009) & 0.712 (-0.046) & 0.758 & 0.774  \\
 MS-SSIM(+) & 0.762 (-0.149) & 0.897 (-0.014) & 0.884 (-0.027) & 0.911 & 0.918 \\
 RMSE(-) & 29.379 (+12.231) & 21.285 (+4.137) & 19.834 (+2.686) & 17.148 & 16.592 \\
 LPIPS(-) & 0.153 (+0.07) & 0.086 (+0.003) & 0.099 (+0.016) & 0.083 & 0.078 \\
 FID(-) & 32.093 (+17.991) & 19.780 (+5.678) & 16.971 (+2.689) & 14.102 & 18.511 \\
\hline
\hline
\end{tabular}}
\caption{Ablation study of our architecture components using CelebA-HQ with the image size of $128 \times 128$. `w/o fusion feedback' denotes the architecture without using the \textit{Fusion Feedback Network}, `w/o attention' denotes the architecture without using the \textit{contextual attention module}, and `w/o self-distill loss' signif{ies} the model trained without using the \textit{self-distillation loss}. \textbf{FULL} is the architecture using all of the proposed components. {Note that all of our models do not use the ground truth semantics but the predicted semantic maps from the semantic map predictor except for the last column (G.T. Semantics).}}
\label{tab:ablation}
    
\end{center}
\end{table*}

\subsection{Comparison {with} baseline models}
We compared our \textit{SGIN} with various image-inpainting models {different in} their types and schemes. For the U-net architecture, we chose Deepfill{-}v2 \cite{yu2019free} {and} Crfill \cite{zeng2021cr}, and for the modulated generator architecture, we chose PsP \cite{richardson2021encoding} {and} E4E \cite{tov2021designing}. We also included SEAN \cite{zhu2020sean} in that it also uses semantic map{s}. and MAT \cite{li2022mat}, the {current} state-of-art (SOTA) inpainting module {which is based on a} transformer model.
For a fair comparison, all of these baseline models are trained with the same \textit{NatOcc} datasets with the same masks obtained from our \textit{Occlusion Detector}, and SEAN is trained with the same semantic map{s} obtained from our \textit{semantic map predictor}, except for MAT whose large computational cost is unaffordable in our devices. 
Alternatively, we made our comparison based on the pretrained CelebA-HQ MAT model uploaded at the author's GitHub repository and used the same masks {as ours}.


\subsection{Quantitative Evaluation}
As some previous papers \cite{yang2019deep,borji2019pros} have pointed out, image generation tasks lack a good quantitative metric for quantitative evaluation. For example, it is possible to have a different but highly plausible reconstructed image from the ground truth but the scores from SSIM or RMSE may fluctuate simply because of its difference from the ground truth. In the light of this, we employed six metrics that can shed light on the different aspects of the quality of reconstruction; PSNR, SSIM \cite{wang2004image}, MS-SSIM \cite{wang2003multiscale}, RMSE, LPIPS \cite{zhang2018unreasonable}, and FID \cite{heusel2017gans}. We evaluated the {average} scores for all of the validation samples. 

Note that U-net generators and the SOTA model MAT conventionally copy-and-paste the rest of the unmasked region from the original image unlike modulated generators that try to reconstruct the whole image. This means that when loss is calculated by comparing the output with its ground truth, they enjoy unfair advantange in the loss score as the unmasked region of the their output is always the same as the ground truth. 
Therefore, we measured two different scores; one of which is measured from the whole generated image and the other measured with the masked region only, where the unmasked region has pixel value of 0. {In Table~\ref{tab:model_compare}}, each cell contains two values, where the left {is} calculated with the whole image, while the {other} at the right {is} calculated only with the masked region. 
As the table shows, our model scored less in the whole image compared to the models which copy-and-paste the unmasked region, but for the scores {obtained only from} the masked region, {ours are} the best in five metrics out of six {(red)}. 
{Among the modulated generators, ours show the best performance for all the six metrics (blue).}

\subsection{Qualitative Evaluation}

Figure \ref{fig:qualitative} shows image inpainting results from various models against the four types of occluding masks of common objects (Object mask), human hands (Hand mask), rectangular masks occluding half of the image (Half mask), and rectangular masks at the center taking 70\% of the image (Center mask). Two U-net generators (DeepFill{-}v2 and Crfill) still suffer from some {noticeable} glitches in their reconstructed images for the masks seen during training (Object and Hand mask) and can produce no acceptable reconstructions for other types of masks, which points out the known problem in the U-net generator's generalization capability. Note that despite this, they scored reasonably high for both whole and mask-only analysis in all of the six quantitative evaluation {metrics}. For modulated generator models (PsP and E4E) and SEAN, their output image is very natural and works {relatively well for} all four types of masks, but fine features like the orientation of the eyes and other details such as the background are not retained, {which} is also one of the reasons why these models score low at quantitative analysis of the whole image. Our model can faithfully regenerate the {occluded} regions of the face and match the overall performance of MAT, even though ours require far less computational resources.


\begin{figure}[h!]
\centering
\includegraphics[width=1.\linewidth]{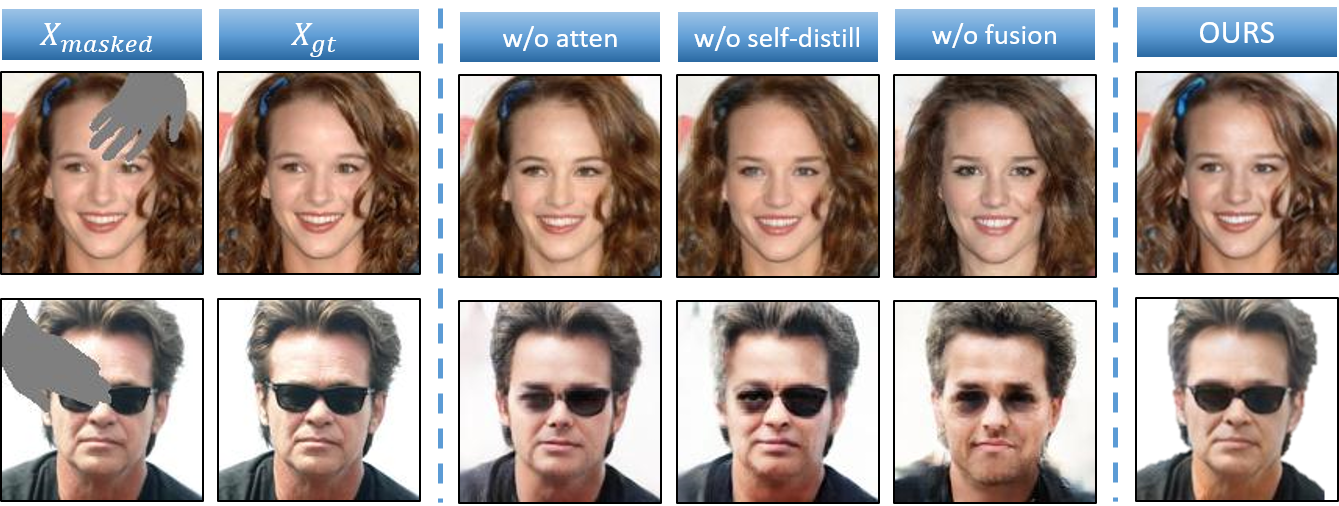}
\caption{\textbf{Qualitative Analysis of the ablation study} `w/o atten' indicates the absence of \textit{attention module} in the style encoder, `w/o self-distill' indicates the absence of \textit{self-distillation loss}, and `w/o fusion' denotes the absence of the \textit{Fusion Feedback module}. {Ours is our full model.} }
\label{fig:ablation}
\end{figure}

\subsection{Ablation Study}

Three of the key introductions of our SGIN architecture from the conventional inpainting networks are {1)} the attention module in the encoder, {2)} the self-distillation loss, and {3)} the fusion feedback module. We ablated each of these changes to assess their impact on the network with the same metrics from the quantitative evaluation. {As Table~\ref{tab:ablation} shows}, the absence of {the} Fusion Feedback Module was most aggravating, and it can be clearly seen from {Fig.~\ref{fig:ablation} in that} the color of the reconstructed images blurs and other fine details are missing. The ablation of both {the} self-distillation loss and the attention module also worsens every score of the metrics. Additionally, we also included scores using the ground truth semantic label in Table~\ref{tab:ablation}. There was no difference between using the ground truth semantic labels and using the predicted labels from our semantic predictor, implying that the effectiveness of our pre-trained semantic predictor. 

\section{Various Applications of SGIN}
Lots of conventional image inpainting models copy-and-paste the unmasked region so the generator only learns how to construct the masked region. It deters the generator from flexible reconstruction of an image, so the model can only control the masked region. 
Different from the conventional models, our inpainting module learns to construct the image as a whole. Therefore, it has {a} control over not only the masked region, but also the unmasked regions. Our model has universal applicability in terms of broad controllability. 

\subsection{Semantic map manipulation}
\begin{figure}[t!] 
\centering
\includegraphics[width=\linewidth]{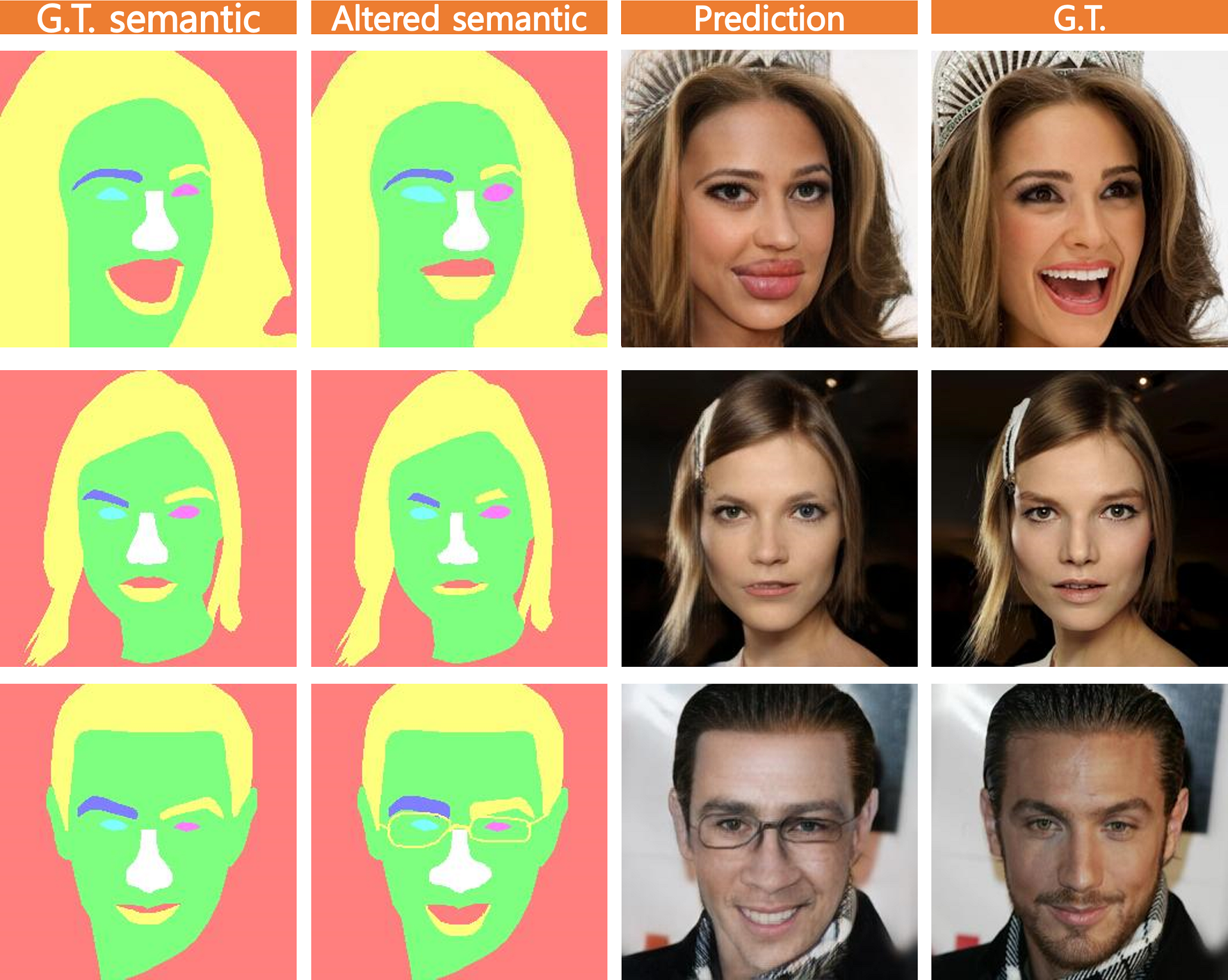}
\caption{\textbf{Semantic map manipulation}}
\label{fig:sem_manipulation}
\end{figure}

Our SGIN module receives two separate inputs, the semantic map and the latent vectors encoded for style and spatial information. The output is constructed in accordance with its semantic map, and by changing it, the user can manipulate the outcome of reconstruction as the user desires. Fig.~\ref{fig:sem_manipulation} shows {some examples} of semantic map manipulation. Our model is capable of creating extra facial features such as glasses and lifted eyes, and altering facial features to have larger eyes, closed lips, shortened eyebrows and opened mouth. 

\subsection{Style Swapping}
\begin{figure}[t!]
\centering
\includegraphics[width=\linewidth]{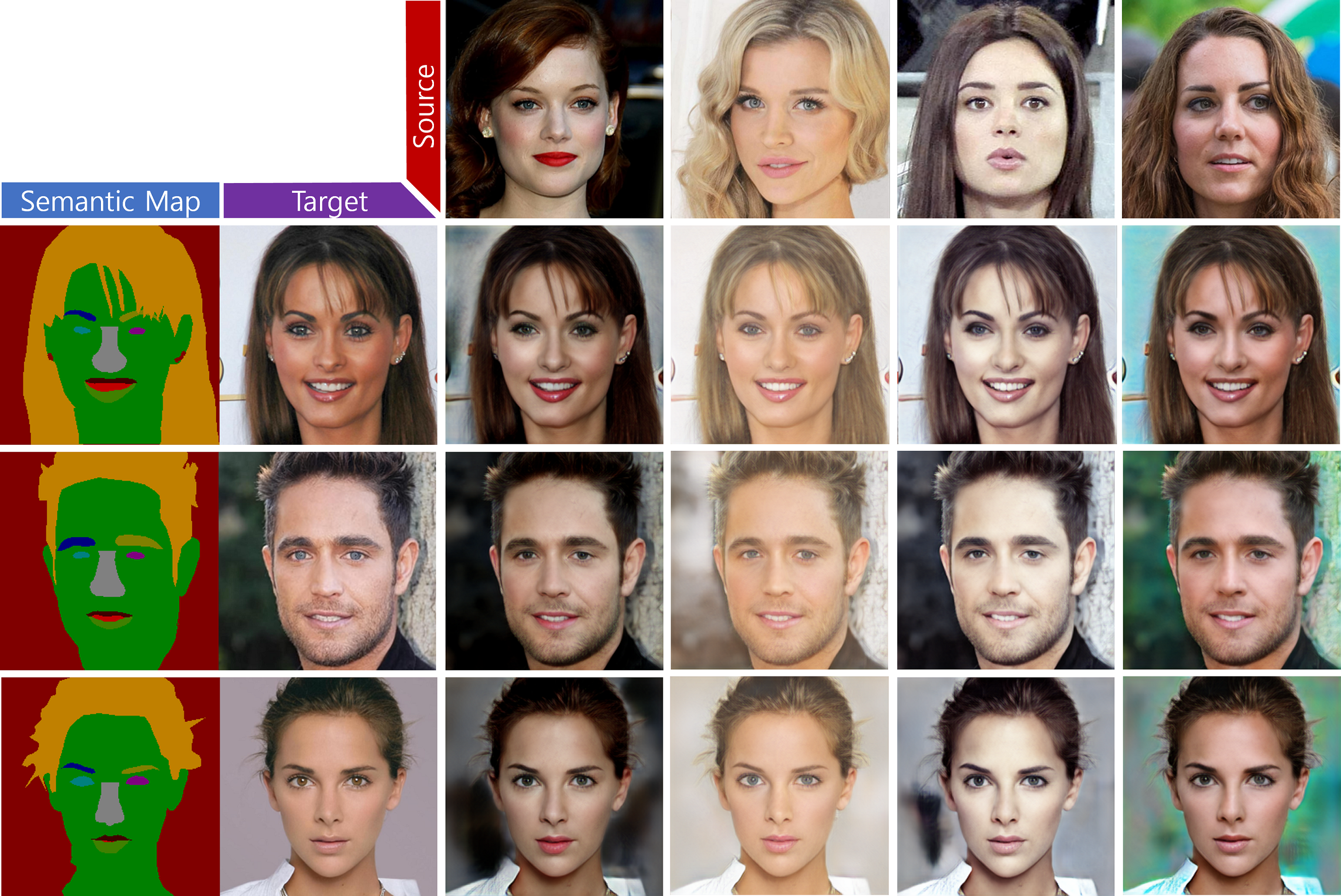}
\caption{\textbf{Examples of Style Swapping}}
\label{fig:style}
\end{figure}

The use of SEAN at the normalization blocks enables simultaneous control over both the spatial and style information. By referring to the style from {an}other image or using {a} user-defined style, it is possible to change the outcome of the reconstruction. Fig.~\ref{fig:style} shows the examples of style swapping, done by swapping the extracted style latent vectors from the \textit{style encoder} of the target image and the source image. 

\begin{figure}[t!]
\centering
\includegraphics[width=\linewidth]{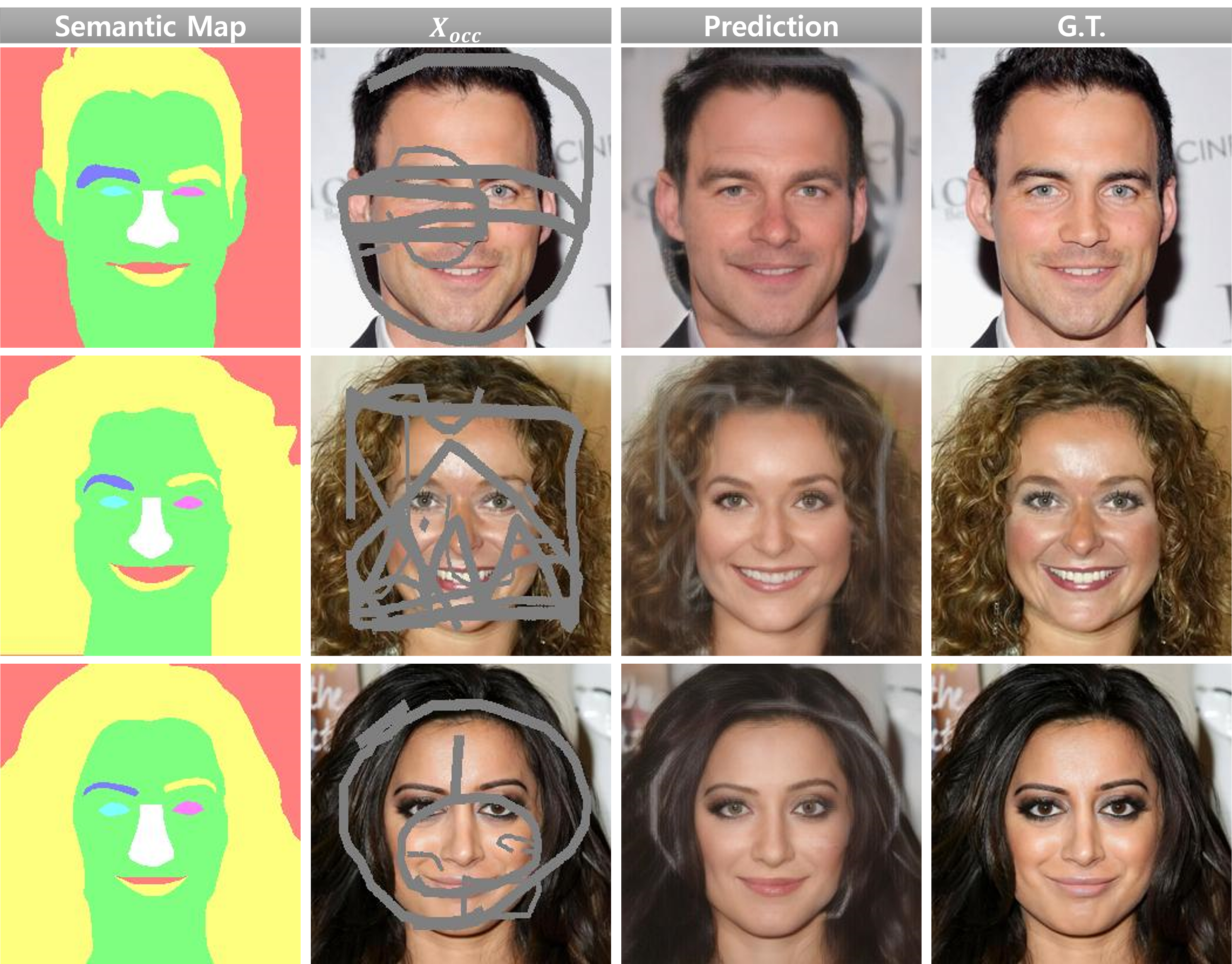}
\caption{\textbf{Failure cases}}
\label{fig:failure}
\end{figure}

\section{Discussions}
\subsection{Failure Cases}

Although our model is relatively robust to various mask types, we found some cases where our model leaves unnatural artifacts. Fig.~\ref{fig:failure} shows various failure cases, most of {which} are irregular masks with thin lines. Our model prediction seems {to well-construct} the region of the face, but {for} other regions such as hair or background, it leaves the traces of the mask. This is mainly due to the characteristics of our model, that it preserves the details in the unmasked region. Therefore, it regards the thin masks as part of the image, especially when the thin mask is on the non-facial region. Our model's shortcoming could {have been} relieved {if it were} trained with more various mask examples.

\section{Conclusion}

We demonstrated that the integration of semantic map, self-distillation loss, and fusion feedback module not only retains the mask-independence of the modulated generator, but also guides it to reconstruct the output image in high consistency with the unmasked image. As our model uses a semantic map of a human face, it also enables users to directly control the reconstructed image by feeding the network with the facial semantic map that includes desired features. We expect that more generalized architecture with the same rationale will usher the advent of an inpainting model with broad applicability. 

\bibliographystyle{splncs04}
\bibliography{aaai23}
\end{document}